# Growth Patterns of Inference

**Abhishek Sharma**

3840 Far West Blvd
Austin, TX 78731
Abhishek81@gmail.com

**Abstract**

What properties of a first-order search space support/hinder inference? What kinds of facts would be most effective to learn? Answering these questions is essential for understanding the dynamics of deductive reasoning and creating large-scale knowledge-based learning systems that support efficient inference. We address these questions by developing a model of how the distribution of ground facts affects inference performance in search spaces. Experiments suggest that uniform search spaces are suitable for larger KBs whereas search spaces with skewed degree distribution show better performance in smaller KBs. A sharp transition in Q/A performance is seen in some cases, suggesting that analysis of the structure of search spaces with existing knowledge should be used to guide the acquisition of new ground facts in learning systems.

## Introduction and Motivation

In recent years, there has been considerable interest in Learning by Reading [Barker et al 2007; Forbus et al 2007, Mulkar et al 2007] and Machine Reading [Etzioni et al 2005; Carlson et al 2010] systems. Such systems are already good at accumulating large bodies of ground facts (although learning general quantified knowledge is currently still beyond the state of the art). But what ground facts should they be learning, to support deductive reasoning? Ideally, new facts should lead to improvements in deductive Q/A coverage, i.e. more questions are answered. Will the rate of performance improvement always be uniform, or will there be "phase changes"? Understanding the dynamics of inference is important to answering these questions, which in turn are important for making self-guiding learning systems.

Our analysis draws upon ideas from network analysis, where the networks are the AND/OR connection graph of a set of first-order Horn axioms. By analogy to epidemiological models, we explore diffusion of inference in the network, i.e. how does coverage of queries increase as new ground facts are learned. Cascade conditions correspond to when inference becomes easy, i.e. increased coverage. Here we argue that some useful insights about growth patterns of inference can be derived from simple features of search spaces. We focus on three parameters: The first, $\alpha$, associated with each node, represents the contribution of each node in answering a set of questions. Parameters $k$ and $\beta$ represent the connectivity of the graph. We study Q/A performance for different values of these parameters, including several sizes of KB contents, to simulate the impact of learning. We found that search spaces with skewed degree distribution lead to better Q/A performance in smaller KBs, whereas in larger KBs more uniform search spaces perform better. In some cases, as $\alpha$ increases, the percolation of inference shows a significant and abrupt change. A degenerate case, in which the effect of ground facts "dies down" and expected improvements in Q/A performance are not observed due to mismatch of expectations and ground facts, is also seen.

The rest of this paper is organized as follows: We start by summarizing related work and the conventions we assume for representation and reasoning. A detailed description of the diffusion model and experimental results are described next. In the final section, we summarize our main conclusions.

## Related Work

In social sciences, there has been significant interest in models of different kinds of cascades. In these domains, the interest is to study how small initial shocks can cascade to affect or disrupt large systems that have proven stable with respect to similar disturbances in the past [Watts 2002]. The model described here is inspired by work on cascades in random graphs [Watts 2002] and epidemic thresholds in networks [Chakrabarti et al 2008]. In AI, there has been work on viral marketing [Domingos & Richardson 2001] and phase transitions in relational learning [Giordana & Saitta 2000], who uses somewhat similar parameter definitions to ours. However, neither of them addresses deductive reasoning in first-order knowledge bases, as we do.

## Representation and Reasoning

We use conventions from Cyc [Matuszek *et al* 2006] in this paper since that is the major source of knowledge base contents used in our experiments[1]. We summarize the key conventions here. Cyc represents concepts as *collections*. Each collection is a kind or type of thing whose instances share a certain property, attribute, or feature. For example, Cat is the collection of all and only cats. Collections are arranged hierarchically by the `genls` relation. `(genls <sub> <super>)` means that anything that is an instance of <*sub*> is also an instance of <*super*>. For example, `(genls Dog Mammal)` holds. Moreover, `(isa <thing> <collection>)` means that <*thing*> is an instance of collection <*collection*>. Predicates are also arranged in hierarchies. Here, `(genlPreds <s> <g>)` means that the predicate <*g*> is a generalization of <*s*>. For example, `(genlPreds touches near)` means that touching something implies being near to it. The set of `genlPreds` statements, like the `genls` statements, forms a lattice. Here `(argIsa <relation> <n> <col>)` means that to be semantically well-formed, anything given as the <*n*>th argument to <*relation*> must be an instance of <*col*>. That is, (<*relation*>……<*arg-n*> …) is semantically well-formed only if `(isa <arg-n> <col>)` holds. For example, `(argIsa mother 1 Animal)` holds.

Learning by Reading systems typically use a Q/A system to examine what the system has learned. For example, Learning Reader used a parameterized question template scheme [Cohen *et al*, 1998] to ask ten types of questions. The templates were: (1) Who was the actor of <*Event*>?, (2) Where did <*Event*> occur?, (3) Where might <*Person*> be?, (4) What are the goals of <*Person*>?, (5) What are the consequences of <*Event*>?, (6) When did <*Event*> occur?, (7) Who was affected by the <*Event*>?, (8) Who is acquainted with (or knows) <*Person*>?, (9) Why did <*Event*> occur?, and (10) Where is <*GeographicalRegion*>? In each template, the parameter (e.g., <*Person*>) indicates the kind of thing for which the question makes sense (specifically, a collection in the Cyc ontology). Each template has a single open variable (e.g. the actor in Q1, a location in Q2, etc.) which must be inferred to provide an answer. We use these questions in our experiments below, to provide realistic test of reasoning.

When answering a parameterized question, each template expands into a set of formal queries, all of which are attempted in order to answer the original question. Our FIRE reasoning system uses backchaining over Horn clauses with an LTMS [Forbus & de Kleer 93]. We limit inference to Horn clauses for tractability. To construct a set of axioms for our experiments, we generate first-order Horn clauses from the general clauses in the ResearchCyc knowledge base. Starting with the queries for the 10 parameterized question types, we generate all Horn clauses relevant for answering them and continue this process recursively for their children until depth 10. This leads to a set of 7,330 first-order axioms. Some heuristics for improving inference are discussed in [Sharma et al 2016, Sharma et al 2019, Sharma and Forbus 2013, Sharma & Forbus 2010, Sharma & Goolsbey 2017]. In the next section, we use different subsets of this axiom set for studying the properties of search spaces.

[1] We use knowledge extracted from the ResearchCyc knowledge base with our own reasoning system, instead of using Cycorp's reasoning system.

## A Model for Spread of Inference

How does the possibility of inference cascades depend on the size of KB and the network of interconnections? In an inference cascade, the effects of new ground facts reach the target queries quickly, and a small number of new facts should lead to disproportionate effects on the final Q/A performance.

To study these issues, it is useful to view a learning system as a *dynamical system*, where the *state* of the system is given by a set of parameters. In what follows, we define some parameters which are useful for describing a knowledge-based learning system. Then we report experimental results for different values of these parameters. It follows that the aim of a meta-reasoning module should be to identify more desirable states of such a system and use this information for guidance

Now, we describe our model of inference propagation. The graph G is cycle-free AND/OR graph generated during the backward chaining of the axiom-set construction process, with N being the set of nodes in this graph. We explore variations in the structure of the search space by choosing subsets of N, say M, and keeping only the edges of G which directly connect nodes in M. In order to consider the space of inferences that could be done with a search space, we define Q to be the set of specific parameterized questions which could be asked for all 10 of the questions defined above. That is, the variable representing the parameter (e.g. <*Event*>) is bound t all possible entities in the KB of that type, whereas the other parameter in each query remains open, to be solved for. For every node m, *depth(m)* represents its depth in G. We can now define α as follows:

$$\alpha = \frac{1}{|N|} \sum_{m \in M} \frac{Solutions(m)}{|Q| * (depth(m)+1)}$$

*Solutions(m)* represents the number of answers returned by the node *m* on its own (i.e., purely by ground fact retrieval and not by using axioms). α represents the average contribution of each node towards answering the set of queries. Depth of nodes has been used to weigh the contribution of nodes because solutions closer to the root node are more likely to percolate up due to fewer unification problems. Another factor which plays an important role in determining the diffusion of inference in search spaces is their connectivity. The larger the degree of nodes, the more likely it is to return a solution by using answers from its neighbors. Moreover, since different degree distributions lead to significant differences in the properties of dynamics of networks, we study how they affect the rate of percolation of inference in the search space.

**Model 1**: We start with the set of 7,330 axioms discussed above. We begin with the target queries and choose *k* children for them at random. If the node has less than *k* children, we choose all of its children. This is done recursively for the children nodes until depth 10. In other words, by construction, the degree of all nodes is less than or equal to *k*. For example, let us consider the search space shown in Figure 2. Then Figure 3 and 4 show examples of search spaces which could be generated from this model when *k* is 2. (Colored nodes in these figures represent selected nodes.)

**Model 2**: We start with a set of 7,330 axioms discussed above. We begin with target queries and choose β% children for them at random. This is done recursively for the children nodes until depth 10. For example, if β=50, then Figure 5 shows an example of search space generated from this model.

**Figure 1: Description of two models**

In this work, we study the properties of two types of degree distributions: (i) Uniform distribution and (ii) Skewed distribution (i.e., resembling scale-free). To generate search spaces from these distributions, we do the following. We start with the axiom set defined in the previous section. By selecting different subsets of these axioms, we can understand how the structure of search space affects inference. As we select different subsets, we are accessing different regions of KB. Since facts are not uniformly distributed in the KB, α for these axiom subsets varies significantly. Moreover, axioms in ResearchCyc KB have a skewed degree distribution. In other words, most predicates have very few children in the query graph, whereas there are many axioms for proving a small number of predicates. See Figure 2 for a simplified example. (For simplicity, only OR nodes have been shown.) A description of two methods for generating these search spaces is shown in Figure 1. Model 1 makes the search space more uniform by limiting the maximum number of children to *k*. On the other hand, Model 2 preserves the skewed degree distribution by selecting β% children of each node. Which of these models is better for better Q/A performance? Moreover, how does the performance depend on *k* and β? We believe that these parameters (i.e., α, β and *k*), play an important role in understanding the dynamics of inference and the possibility of inference cascades in deductive search spaces. To understand the performance differences between these search spaces, we generated a number of instances of these models. For Model 1, we generated axiom sets for 2≤*k*≤7. Similarly, for Model 2, we generated axiom sets where β ε {10, 15, 20, 30, 40, 50}. For each value of *k* and β, at least seven sets of search spaces were generated[2].

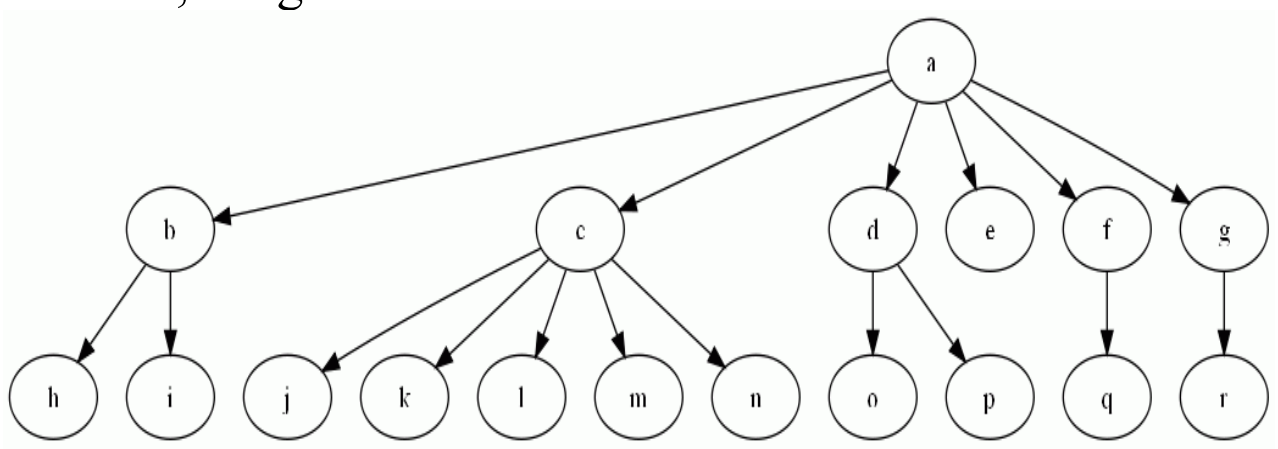

Figure 2: An example of a search space with skewed degree distribution.

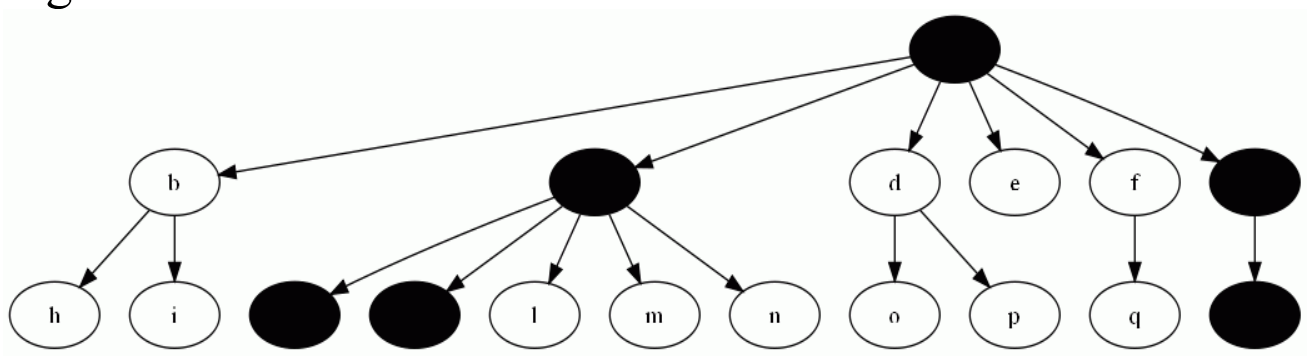

Figure 3: An example of Model 1 search space. Here *k*=2.

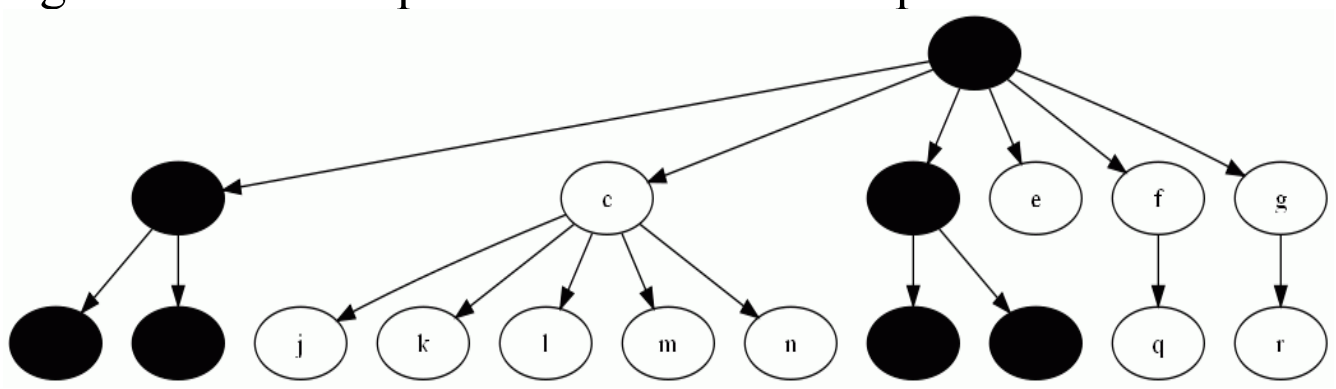

Figure 4: Another example of Model 1 search space. Here *k*=2.

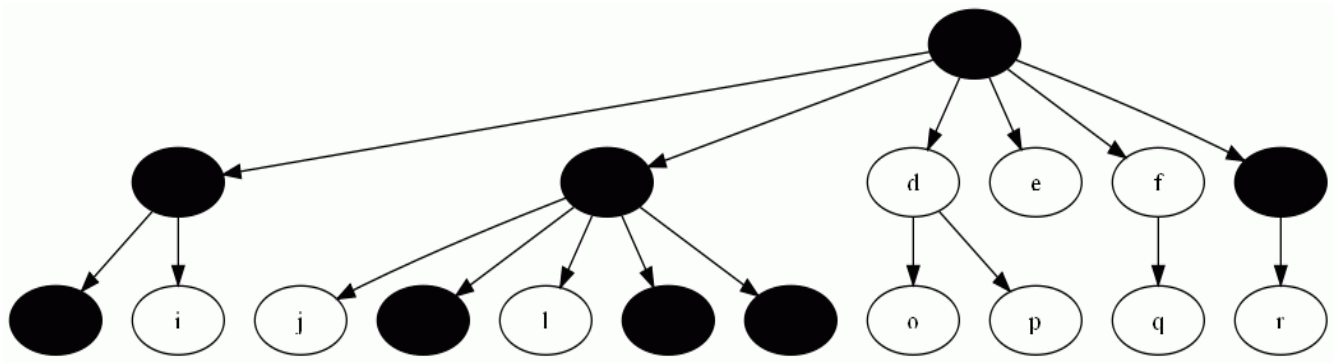

Figure 5: An example of Model 2 search space obtained from the search space shown in Figure 2. (β=50).

As learning systems gather thousands of facts from reading and other sources, the size of their KBs would grow. To model and understand the effects of increasing KB size on the performance of learning systems, we use the inverse ablation model [Sharma & Forbus 2010]. The basic idea is to take the contents of a large KB (here, ResearchCyc KB) and create a small KB by ablation and incrementally re-add facts, collecting snapshots of reasoning performance of the system. We use this method to generate two KBs of size 5,180 and 165,992 facts respectively. We also use the original ResearchCyc KB which contains 491,091 facts[3]. Therefore, the model presented here has been evaluated on these three KBs. In what follows, they are referred to as $KB_1$, $KB_2$ and $KB_3$ respectively.

[2] In some cases, more than 7 sets were generated for better understanding the patterns.

It is obvious that as α increases, we should be able to answer more questions. However, we found that in about 28% of all cases, search spaces with high α did not lead to expected high performance. This abnormality is mainly seen when minimal unification takes place at a small number of nodes which lie in all paths from the leaves to the root nodes. An example of this phenomenon is shown in Figure 6. We see how inference propagates from the leaves (depth 9) to the root nodes (depth 0). We observe that for the degenerate case, about 120,000 facts are inferred at depth 5. However, this was reduced to almost zero answers at depth 3. The other search space accesses less ground fact rich regions (about 27,000 facts at depth 2), but manages to produce more than 10,000 answers at the root node. This shows that small mismatch between the expectations of axioms and ground facts can lead to serious problems in inference propagation. In this particular degenerate case, no unification took place at the node which joined the ground fact rich regions to the root node. Although the models of selecting axioms described here are admittedly simple, it is surprising that 28% of all axiom-sets generated by them have such serious problems. This implies that knowledge acquisition process has to be informed of the expectations of inference chains. In the absence of such a process, the effects of thousands of facts learnt from different sources would degenerate, and the results of learning would not show up in the final Q/A performance.

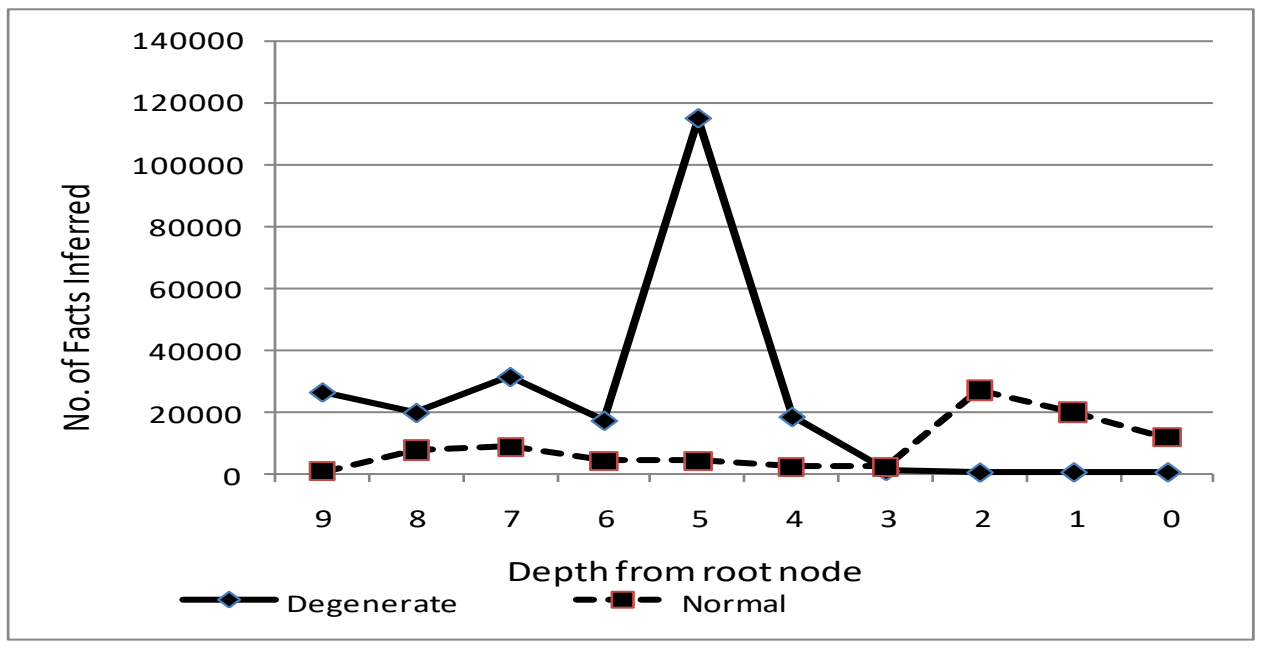

**Figure 6: An example of a degenerate case**

## Experimental Results

In this section, we study how α, β and k affect the dynamics of Q/A performance. Recall the set of 10 questions discussed before. All questions which satisfy the constraints of these templates were generated. $KB_1$, $KB_2$ and $KB_3$ led to 5409, 13938 and 36,564 queries respectively. In particular, we would like to answer following questions: (1) How does Q/A performance change as we access those regions of KB which have more facts (i.e., α increases)? (2) What is the nature of Q/A performance as search spaces become denser (i.e., as *k* and β increase) and (3) Under what conditions does inference percolate to a sizeable section of the KB helping us to answer more than a given threshold fraction of questions? (In this work, the threshold is 0.2)

In Figure 7, we observe the average performance of Model 1 search spaces for the three KBs discussed before. We observe that the threshold performance was not reached for any value of *k* in the smallest KB. On the other hand, as *k* increases, performance gradually improves. Moreover, as KB becomes bigger, the threshold is achieved for sparser search spaces. In Figure 8, we see similar trends for Model 2 search spaces. The only significant difference is that search spaces for β>30 attain threshold performance in $KB_1$ as well. Figure 7 and 8 show that larger KBs lead to better Q/A performance even with fewer axioms. It is interesting to note that there is a very small difference in the performance in different KBs for higher values of β in Model 2 search spaces, whereas their performance varies significantly in Model 1 search spaces. This brings us to another interesting question: Which of the two models discussed here lead to better Q/A performance? We note that although higher values of k and β imply higher connectivity, there is no one-to-one correspondence between these parameters. Therefore, to compare these two models, we selected a set of axiom-sets which had same average degree from both models. The average number of answers for these axiom-sets was then measured. The results are shown in Table 1. We see that while uniform search spaces perform better for larger KBs, search spaces with skewed degree distribution perform better in smaller KBs.

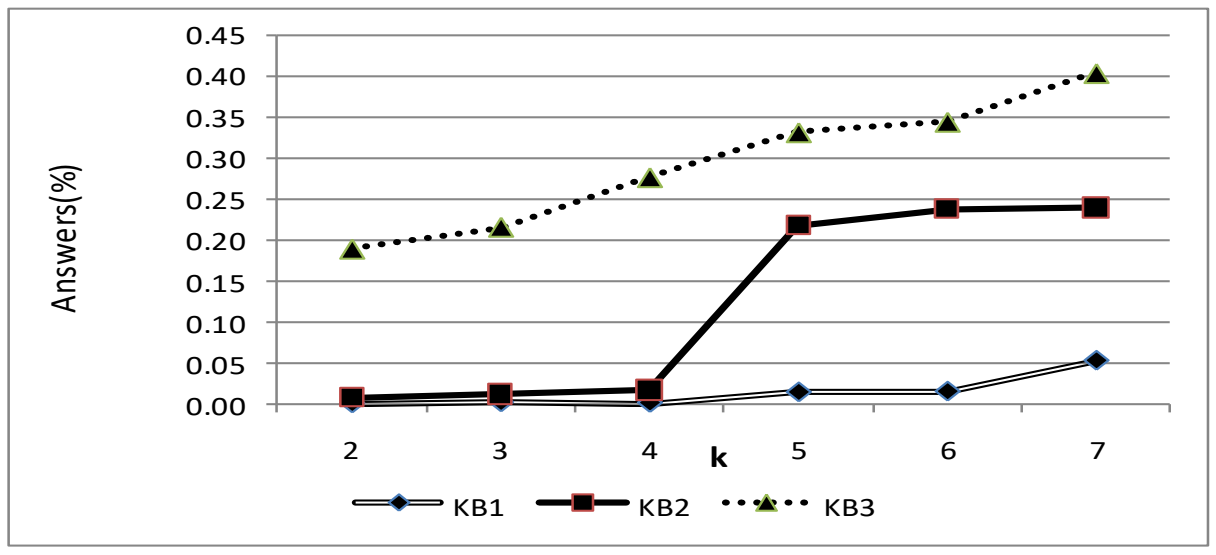

**Figure 7: Q/A performance for Model 1 search spaces**.

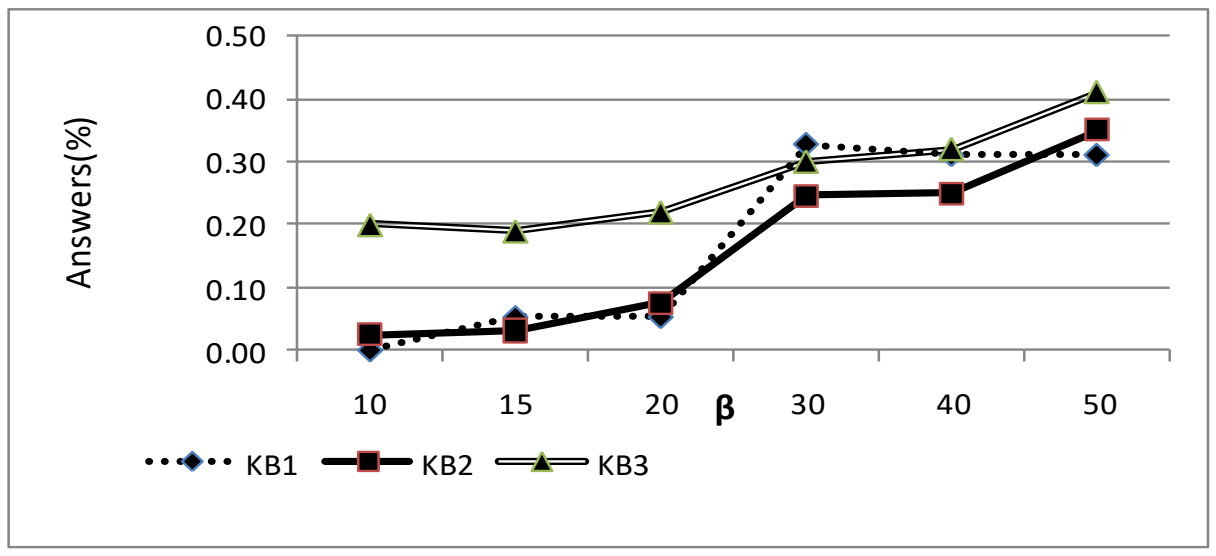

**Figure 8: Q/A performance for Model 2 search spaces**.

[3] These numbers do not include the size of the ontology. The total size of our ResearchCyc KB is 1.2 million facts.

| KB | Model 1 | Model 2 | Change in Model 2 w.r.t. Model 1 |
|---|---|---|---|
| $KB_1$ | 33 | 50 | +51.5% |
| $KB_2$ | 840 | 213 | -74.6% |
| $KB_3$ | 10176 | 9512 | -6.5% |

**Table 1: Average number of answers for two models**

Next, we discuss the change in Q/A performance as α increases. How does the performance change as we access those regions of KB which are richer in facts? Figure 9 shows a clear transition between a phase in which almost no answers are inferred to a high inference phase. However, for $KB_1$, such critical transition was observed only for denser graphs in Model 1 search spaces (i.e., k=7). In other cases, the number of answers inferred remained very low. Figure 10 shows similar transition for β=30 and 50 in Model 2 search spaces.

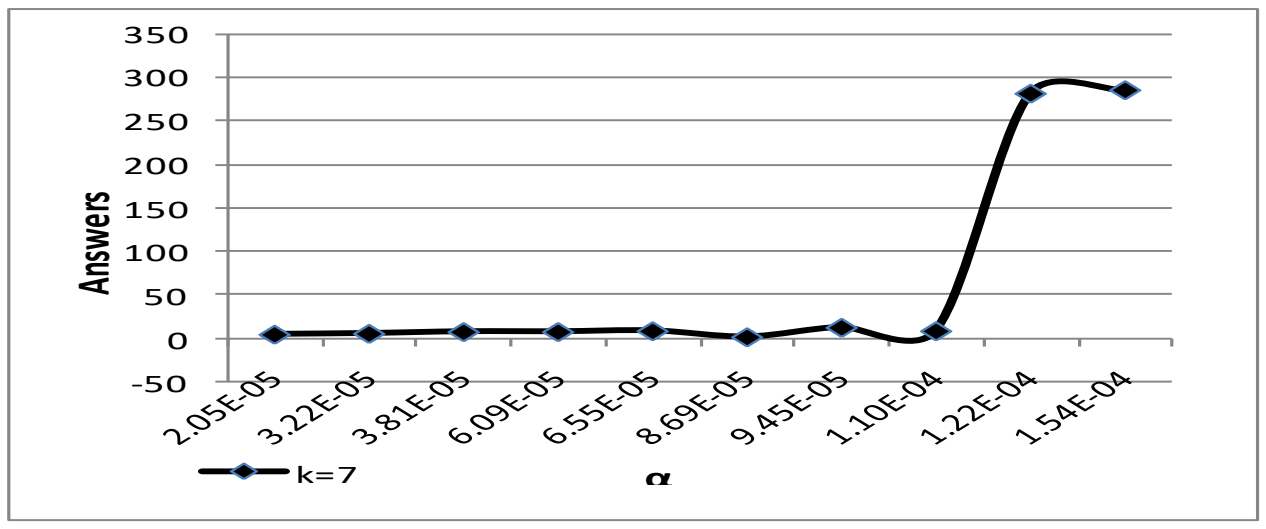

**Figure 9: Q/A performance for Model 1 search space for $KB_1$, k=7.**

In $KB_2$, a quick transition from a low inference to high inference is seen in more cases. For Model 1 search spaces, such quick increase in seen when k is 5 or 6 (see Figure 11 for the case when k is 6). Similarly, Model 2 search spaces also show two distinct phases of inference (see Figure 12 and 13). Figure 14 shows examples of critical transitions for $KB_3$.

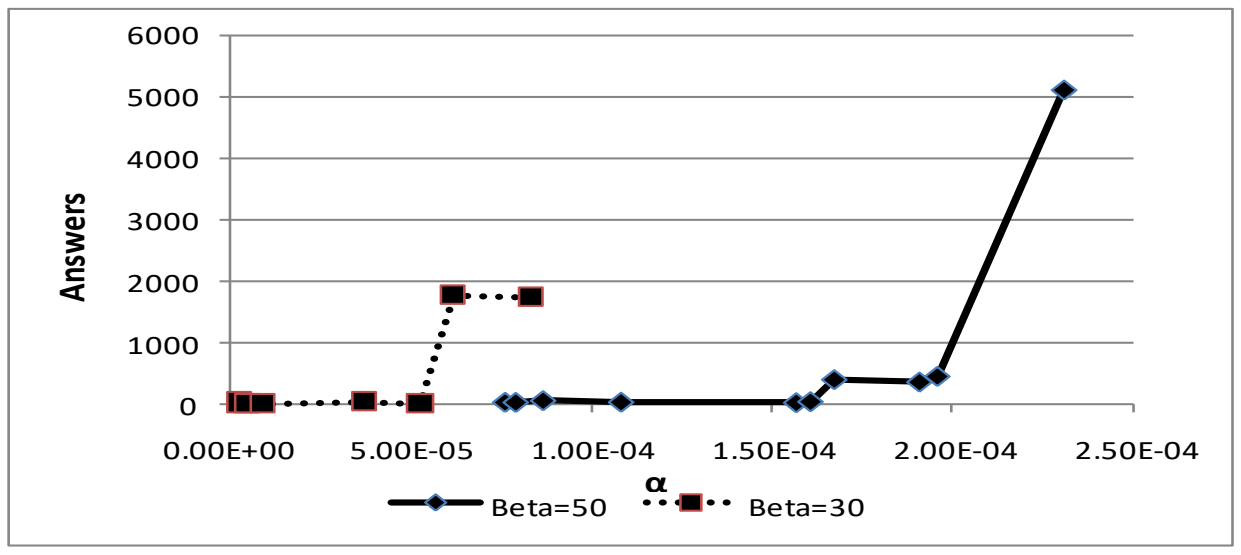

**Figure 10: Q/A performance for Model 2 search space in $KB_1$.**

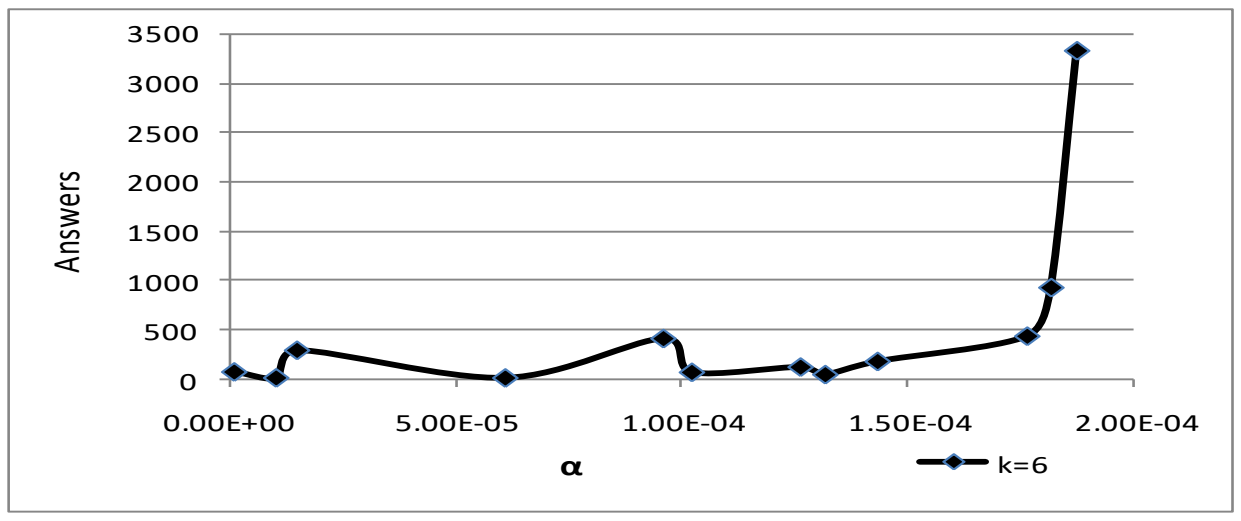

**Figure 11: Q/A performance for a Model 1 search space in $KB_2$.**

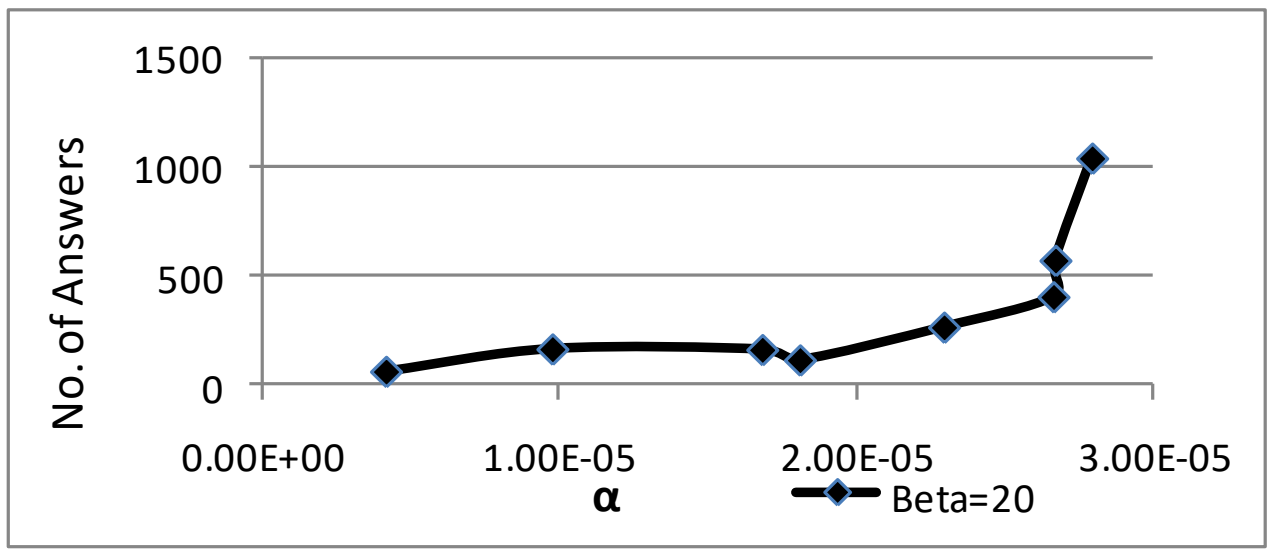

Figure 12: Q/A performance for Model 2 search space in $KB_2$.

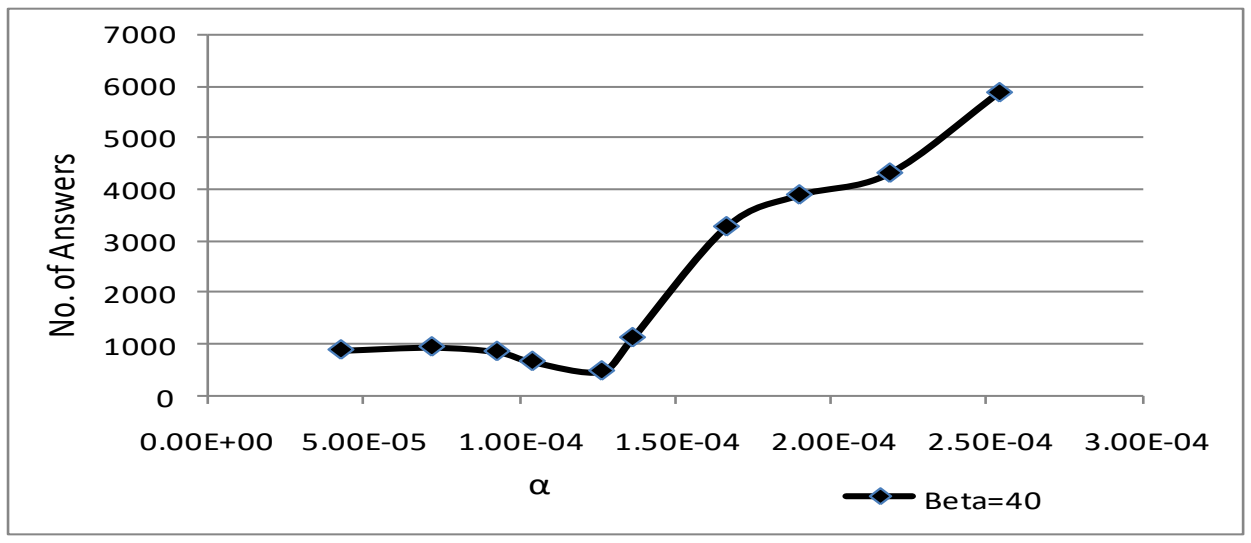

**Figure 13: Q/A performance for Model 2 search space in $KB_2$.**

In our experiments, about 36% of all search spaces did show a sharp transition discussed above. The results imply that same amount of learning effort can lead to significantly different performance depending on the state of the system. For example, a small addition of facts when the value of α is low will lead to only modest improvement in Q/A performance. On the other hand, when the system is close to transition point to fast inference, a small amount of learning can provide disproportionate payoff in terms of performance.

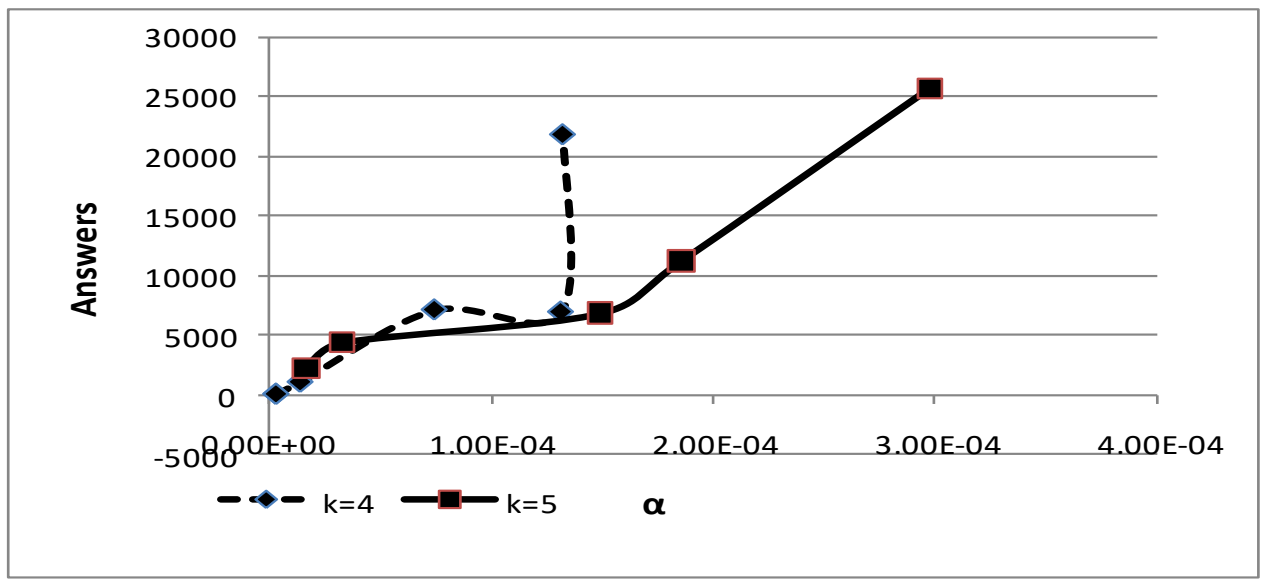

**Figure 14: Q/A performance for Model 1 search space for $KB_3$, k=4 and k=5.**

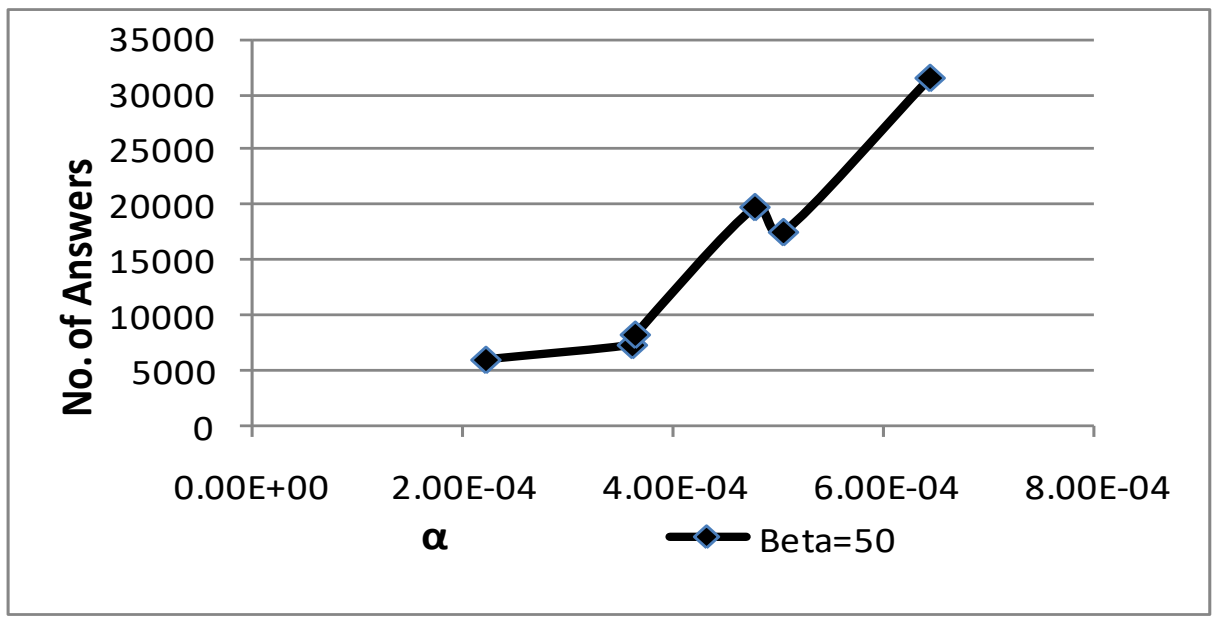

**Figure 15: No sharp transition for this Model 2 search space in $KB_3$.**

However, in remaining 36% of all cases, no discernible change in the Q/A performance was observed (see Fig 15 for an example). In some cases, the spread of inference was limited by the sparseness of the search space. In other cases, larger KBs already showed reasonably high Q/A performance for low values of α, and further improvements were limited by low density of facts.

## Conclusions

As large-scale learning systems mature, there will be a need to steer their learning towards states which lead to progressively better Q/A performance. The study of the structure of search spaces and knowledge, and dynamics of inference are important for attaining this goal. We have proposed and analyzed a model in which simple features of search spaces and knowledge base are used to study the growth patterns of inference. We have reported results for two types of degree distributions and three KBs. The propagation of inference is much less in smaller KBs. Search spaces with uniform degree distributions perform better in larger KBs, whereas relatively skewed degree distributions are more suitable for smaller KBs. Small but critical mismatch between the expectations of axioms and facts in the KB, which lead to almost zero inferences, were observed in 28% of axiom sets generated from the models discussed here. In 36% of all cases, a critical transition between a low-inference to a high-inference region was observed. Next generation learning systems should be cognizant of these properties and the knowledge acquisition cycle should be pro-active in guiding the system towards high-inference states. It is hoped that the introduction and study of this model will stimulate further research into understanding the properties of first-order inference and its dependence on the distribution of facts in the KB.